\documentclass[conference]{IEEEtran}
\usepackage{multirow}
\usepackage{times}
\usepackage{url}
\usepackage{graphicx,epstopdf}
\usepackage{latexsym}
\usepackage[tight]{subfigure}
\usepackage{algorithm}
\usepackage{algorithmic}
\usepackage{amsmath} 
\usepackage{amssymb}  
\usepackage{cases}
\usepackage{float}
\usepackage{array}
\newcolumntype{C}[1]{>{\centering}p{#1}}

\newcommand{\tabincell}[2]{\begin{tabular}{@{}#1@{}}#2\end{tabular}}

\ifCLASSINFOpdf
\else
\fi
\begin{document}
%
\title{Large Margin Nearest Neighbor Embedding for Knowledge Representation}

\author{
%
%

Miao Fan$^{\dagger,\ddagger,*}$, Qiang Zhou$^{\dagger}$, Thomas Fang Zheng$^{\dagger}$ and Ralph Grishman$^{\ddagger}$\\
$^{\dagger}$ CSLT, Division of Technical Innovation and Development,\\ Tsinghua National Laboratory for Information Science and Technology,\\ Tsinghua University, Beijing, 100084, China.\\
$^{\ddagger}$Proteus Group, New York University, NY, 10003, U.S.A.\\
{\tt $^*$fanmiao.cslt.thu@gmail.com}
}

%


\maketitle

\begin{abstract}
Traditional way of storing facts in triplets ({\it head\_entity, relation, tail\_entity}), abbreviated as ({\it h, r, t}), makes the knowledge intuitively displayed and easily acquired by mankind, but hardly computed or even reasoned by AI machines. Inspired by the success in applying {\it Distributed Representations} to AI-related fields, recent studies expect to represent each entity and relation with a unique low-dimensional embedding, which is different from the symbolic and atomic framework of displaying knowledge in triplets. In this way, the knowledge computing and reasoning can be essentially facilitated by means of a simple {\it vector calculation}, i.e. ${\bf h} + {\bf r} \approx {\bf t}$. We thus contribute an effective model to learn better embeddings satisfying the formula by pulling the positive tail entities ${\bf t^{+}}$ to get together and close to {\bf h} + {\bf r} ({\it Nearest Neighbor}), and simultaneously pushing the negatives ${\bf t^{-}}$ away from the positives ${\bf t^{+}}$ via keeping a {\it Large Margin}. We also design a corresponding learning algorithm to efficiently find the optimal solution based on {\it Stochastic Gradient Descent} in iterative fashion. Quantitative experiments illustrate that our approach can achieve the state-of-the-art performance, compared with several latest methods on some benchmark datasets for two classical applications, i.e. {\it Link prediction} and {\it Triplet classification}. Moreover, we analyze the parameter complexities among all the evaluated models, and analytical results indicate that our model needs fewer computational resources on outperforming the other methods.
\end{abstract}


%
\IEEEpeerreviewmaketitle

\section{Introduction}
Owing to the long-term efforts on distilling the explosive information on the Web, several large scale Knowledge Bases (KBs), such as WordNet\footnote{http://www.princeton.edu/wordnet} \cite{Miller1995}, OpenCyc\footnote{http://www.cyc.com/platform/opencyc} \cite{Lenat1995Cyc}, YAGO\footnote{www.mpi-inf.mpg.de/yago-naga/yago} \cite{suchanek2007WWW} and Freebase\footnote{http://www.freebase.com} \cite{Bollacker2007}, have already been built. Despite the different domains these KBs serve for, nearly all of them concentrate on enriching {\it entities} and {\it relations}. To facilitate storing, displaying and even retrieving knowledge, we use the highly structured form, i.e. ({\it head\_entity, relation, tail\_entity}), for knowledge representation. Each triplet is called a {\it fact}. So far, some general domain KBs, such as Freebase\footnote{According to the statistics released by the official site, Freebase contains 43 million entities and 1.9 billion triplets for now.} and YAGO\footnote{As of 2012, YAGO2s has knowledge of more than 10 million entities and contains more than 120 million facts about these entities.}, have contained millions of entities and billions of facts, and they should have led a huge leap forward for some canonical AI-related tasks, e.g. {\it Question Answering Systems}. However, it is realised that this symbolic and atomic framework of representing knowledge makes it difficult to be utilized by most of AI-machines, especially of those which are dominated by statistical approaches.

\begin{figure*}
\centering
\includegraphics[width=0.8\textwidth]{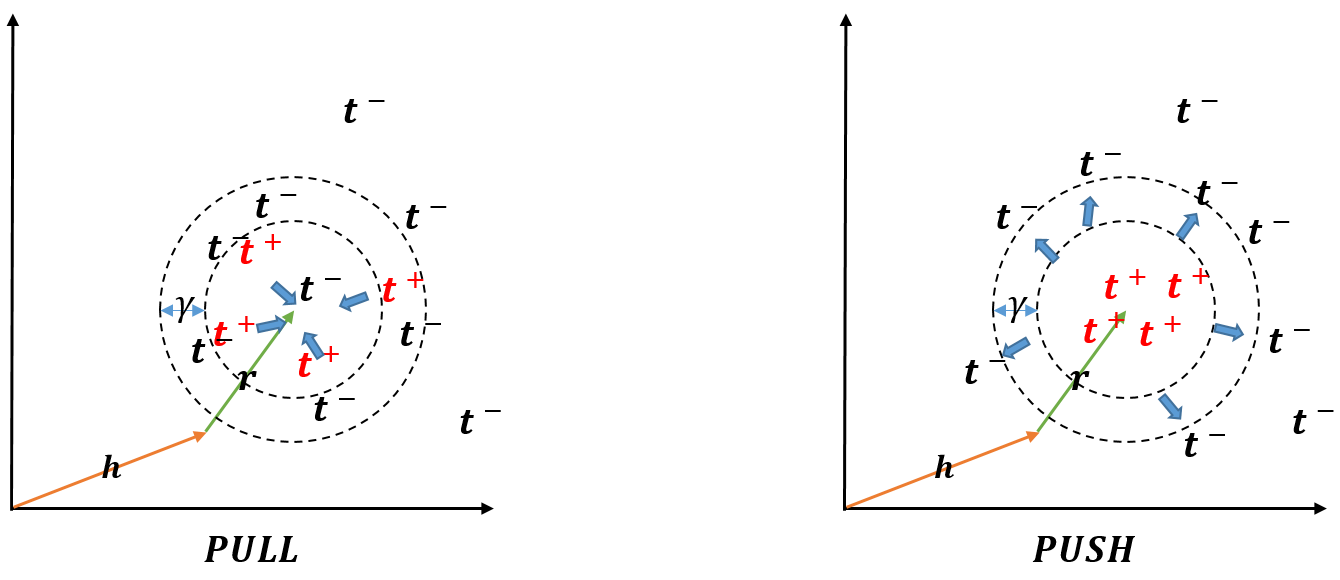}\\

\caption{{\bf LMNNE} has two objects optimized simultaneously: {\it PULL} the positive tail entities (${\bf t}^{+}$) close to ${\bf h + r}$ and {\it PUSH} the negative tail entities (${\bf t}^{-}$) out of the margin ${\gamma}$. In this case, all the entities and relations are embedded into the 2D vector space and we use $L_2$ {\it norm} to measure the distance.}

\end{figure*}

Recently, many AI-related fields, such as {\it Image Classification} \cite{conf/cvpr/CiresanMS12}, {\it Natural Language Understanding} \cite{le2014distributed} and {\it Speech Recognition} \cite{conf/icassp/SchwenkG02}, have made significant progress by means of learning {\it Distributed Representations}.
Taking an example of distributed representations of words applied to {\it Statistical Language Modeling} \cite{journals/jmlr/BengioDVJ03}, it has achieved considerable success by grouping similar words which are close to each other in low-dimensional vector spaces. Moreover, Mikolov et al. \cite{Mikolov2013,conf/naacl/MikolovYZ13} discovered somewhat surprising patterns that the learnt word embeddings, to some extent, implicitly capture syntactic and semantic regularities in language. For example, the result of vector calculation ${\bf v}_{Madrid} - {\bf v}_{Spain} + {\bf v}_{France}$ is closer to ${\bf v}_{Paris}$ than to any other words \cite{mikolov2013distributed}.

Inspired by the idea of word vector calculation, we look forward to making knowledge computable. If we ideally consider the example mentioned above, the most probable reason ${\bf v}_{Madrid} - {\bf v}_{Spain} + {\bf v}_{France} \approx {\bf v}_{Paris}$, is that {\it capital\_city\_of} is the relation between {\it Madrid} and {\it Spain} , and so is {\it Paris} and {\it France}. In other words, we can conclude that:

\begin{itemize}
  \item There are two facts/triplets, i.e. ({\it Madrid}, {\it capital\_city\_of}, {\it Spain}) and ({\it Paris}, {\it capital\_city\_of}, {\it France}).
  \item We also derive the approximate equation that ${\bf v}_{Spain} - {\bf v}_{Madrid} \approx {\bf v}_{France} - {\bf v}_{Paris}$ from ${\bf v}_{Madrid} - {\bf v}_{Spain} + {\bf v}_{France} \approx {\bf v}_{Paris}$.
\end{itemize}

The shared relation {\it capital\_city\_of} may help establish the approximate equation due to certain implicit connection. If we assume that the relation {\it capital\_city\_of} can also be explicitly represented by a vector, i.e. ${\bf v}_{capital\_city\_of}$, the connection will be ${\bf v}_{Spain} - {\bf v}_{Madrid} \approx {\bf v}_{France} - {\bf v}_{Paris} \approx {\bf v}_{capital\_city\_of}$. Therefore, the fact ({\it Madrid}, {\it capital\_city\_of}, {\it Spain}) can be modeled in another way, i.e. ${\bf v}_{Madrid} + {\bf v}_{capital\_city\_of} \approx {\bf v}_{Spain}$.

Generally speaking, this paper explores a better approach on knowledge representation by means of learning a unique low-dimensional embedding for each entity and relation in KBs, so that each fact ({\it head\_entity, relation, tail\_entity}) can be represented by a simple vector calculation ${\bf h} + {\bf r} \approx {\bf t}$. To achieve this goal, we contribute a generic model named {\it Large Margin Nearest Neighbor Embedding} ({\bf LMNNE}). As intuitively shown by Figure 1, {\bf LMNNE} follows the principle of {\it pulling} the positive tail entities ${\bf t}^{+}$ close to ${\bf h + r}$, and simultaneously {\it pushing} the negative tail entities (${\bf t}^{-}$) a large margin ${\gamma}$ away from the positives ${\bf t}^{+}$. The details about model formulation are described in Section 3. Section 4 presents the algorithm that solves {\bf LMNNE} efficiently based on {\it Stochastic Gradient Descent} ({\bf SGD}) in iterative fashion. To prove both effectiveness and efficiency of the proposed model, we conduct quantitative experiments in Section 5, i.e. evaluating the performance of {\it Link prediction} and {\it Triplet classification} on several benchmark datasets. We also perform qualitative analysis via comparing model complexity among all related approaches in Section 6. Results demonstrate that {\bf LMNNE} can achieve the state-of-the-art performance and demand for fewer computational resources compared with many prior arts.

\section{Related Work}
All the related studies work on studying better way of representing a fact/triplet. Usually, they design various scoring functions $f_r(h, t)$ to measure the plausibility of a triplet ($h, r, t$). The lower dissimilarity of the scoring function $f_r(h, t)$ is, the higher compatibility of the triplet ($h, r, t$) will be.

{\bf Unstructured} \cite{Bordes2013a} is a naive model which just exploits the occurrence information of the head and the tail entities without considering the relation between them. It defines a scoring function $||{\bf h}-{\bf t}||$, and this model obviously can not discriminate entity-pairs with different relations. Therefore, {\bf Unstructured} is commonly regarded as the baseline approach.

{\bf Distance Model (SE)} \cite{Bordes2011} uses a pair of matrices $(W_{rh}, W_{rt})$, to characterize a relation $r$. The dissimilarity of a triplet ($h, r, t$) is calculate by the $L_1$ {\it norm} of $||W_{rh}{\bf h} - W_{rt}{\bf t}||$. As pointed out by Socher et al. \cite{Socher2013}, the separating matrices $W_{rh}$ and $W_{rt}$ weaken the capability of capturing correlations between entities and corresponding relations, even though the model takes the relations into consideration.

{\bf Single Layer Model} proposed by Socher et al. \cite{Socher2013} aims at alleviating the shortcomings of {\bf Distance Model} by means of the nonlinearity of a single layer neural network $g(W_{rh}{\bf h} + W_{rt}{\bf t} + {\bf b}_r)$, in which $g = tanh$. Then the linear output layer gives the scoring function: ${\bf u}^T_rg(W_{rh}{\bf h} + W_{rt}{\bf t} + {\bf b}_r)$.

{\bf Bilinear Model} \cite{Sutskever2009,Jenatton2012} is another model that tries to fix the issue of weak interaction between the head and tail entities caused by {\bf Distance Model} with a relation-specific bilinear form: $f_r(h, t) = {\bf h}^TW_r{\bf t}$.

{\bf Neural Tensor Network (NTN)} \cite{Socher2013} proposes a general scoring function:  $f_r(h, t) = {\bf u}^T_rg({\bf h}^TW_r{\bf t}+ W_{rh}{\bf h} + W_{rt}{\bf t} + {\bf b}_r)$, which combines the {\bf Single Layer Model} and the {\bf Bilinear Model}. This model is more expressive as the second-order correlations are also considered into the nonlinear transformation function, but the computational complexity is rather high.

{\bf TransE} \cite{Bordes2013a} is the state-of-the-art method so far. Differing from all the other prior arts, this approach embeds relations into the same vector space of entities by regarding the relation $r$ as a translation from $h$ to $t$, i.e. ${\bf h} + {\bf r} = {\bf t}$. This model works well on the facts with ONE-TO-ONE mapping property, as minimizing the global loss function will impose ${\bf h} + {\bf r}$ equaling to ${\bf t}$. However, the facts with multi-mapping properties, i.e MANY-TO-MANY, MANY-TO-ONE and ONE-TO-MANY, impact the performance of the model. Given a series of facts associated with a ONE-TO-MANY relation $r$, e.g. ${(h, r, t_1), (h, r, t_2), ..., (h, r, t_m)}$, {\bf TransE} tends to represent the embeddings of entities on MANY-side extremely the same with each other and hardly to be discriminated. Moreover, the learning algorithm of {\bf TransE} only considers the randomly built negative triplets, which may bring in bias for embedding entities and relations.

Therefore, we propose a generic model ({\bf LMNNE}) in the subsequent section to tackle the margin-based knowledge embedding problem. This model can fully take advantage of both positive and negative triplets, and in addition, be flexible enough when dealing with the multi-mapping properties.

\section{Proposed Model}
Given a triplet ($h, r, t$), we use the following formula as the scoring function $f_r(h, t)$ to measure its plausibility, i.e.
\begin{equation}
f_r(h, t) = ||h + r - t||,
\end{equation}
where $||\square||$ stands for the generic distance metrics ($L_1 $ {\it norm} or $L_2$ {\it norm}) depending on the model selection.

Ideally, we look forward to learning a unique low-dimensional vector representation for each entity and relation, so that all the triplets in the KB will satisfy the equation ${\bf h + r = t}$. However, this can not be done perfectly because of the subsequent multi-mapping reasons,

\begin{itemize}
  \item Not all entity pairs involve in only one relation. For example, ({\it Barack Obama, president\_of, U.S.A}) and ({\it Barack Obama, born\_in, U.S.A}) are both correct facts in the KB. However, we do not expect that the embedding of the relation {\it president\_of} is the same as the relation {\it born\_in}, since those relations are not semantically related.
  \item Besides the multi-relation property above, we may also face the problem of multiple head entities or tail entities when the other two elements are given. For example, there are at least five correct tail entities, i.e. {\it Comedy film, Adventure film, Science fiction, Fantasy} and {\it Satire}, given ({\it WALL-E, has\_the\_genre, ?}). Likewise, we do not desire that those tail entities share the same embedding.
\end{itemize}

Therefore, we suggest a `soft' way of modeling triplets in KBs. Suppose that $\Delta$ is the set of facts in the KB. For each triplet ($h, r, t$) in $\Delta$, we build a set of reconstructed triplets $\Delta_{(h, r, t)}$ by means of replacing the head or the tail with all the other entities in turn. $\Delta_{(h, r, t)}$ can be divided into two sets, i.e. $\Delta^{+}_{(h, r, t)}$ and $\Delta^{-}_{(h, r, t)}$. $\Delta^{+}_{(h, r, t)}$ is the positive set of triplets reconstructed from ($h, r, t$) as $\Delta^{+}_{(h, r, t)} \subset \Delta$, and $\Delta^{-}_{(h, r, t)}$ is the negative because $\Delta^{-}_{(h, r, t)} \not\subset \Delta$. The intuitive goal of our model is to learn the embeddings of positive tail entities ${\bf t^{+}}$ closer to ${\bf h + r}$ than any other negative embeddings ${\bf t^{-}}$. Therefore, the goal contains two objects, i.e. pulling the positive neighbors near ({\bf NN}) each other while pushing the negatives a large margin ({\bf LM}) away.

Specifically, for a pair of positive triplets, ($h, r, t$) and ($h^{+},r,t^{+}$), we {\it pull} the head or the tail entities together (left panel, Figure 1) by minimizing the loss of variances in distance, i.e.
\begin{equation}
\mathcal{L}_{pull} = \text {Min} \sum_{(h,r,t) \in \Delta} \sum_{(h^{+},r,t^{+})\in \Delta^{+}_{(h,r,t)}} (||h - h^{+}|| + ||t - t^{+}||).
\end{equation}
Simultaneously, we {\it push} the negative head or tail entities away (right panel, Figure 1) via keeping all the negative distances $f_r(h^{-}, t^{-})$ at least one margin $\gamma$ farther than $f_r(h^{+}, t^{+})$. Therefore, the objective function is,
\begin{equation}
\mathcal{L}_{push} = \text {Min} \sum_{(h, r, t) \in \Delta} \sum_{(h^{-}, r, t^{-})\in \Delta^{-}_{(h, r, t)}} [\gamma + f_r(h, t) - f_r(h^{-}, t^{-})]_{+}.
\end{equation}
$[\square]_{+}$ is the hinge loss function equivalent to $max(0, \square)$, which can spot and update the negative triplets that violate the margin constrains.

Finally, we propose the {\it Large Margin Nearest Neighbor Embedding} model which uses $\mu$ to control the trade-off between the pulling $\mathcal{L}_{pull}$ and pushing $\mathcal{L}_{push}$ operations by setting the total loss $\mathcal{L}$ as follows,
\begin{equation}
\mathcal{L} = \text{Min} ~~\mu \mathcal{L}_{pull} + (1 - \mu)\mathcal{L}_{push}.
\end{equation}

\section{Learning Algorithm}
To efficiently search the optimal solution of {\bf LMNNE}, we use {\it Stochastic Gradient Descent} (SGD) to update the embeddings of entities and relations in iterative fashion. As shown in Algorithm 1, we firstly initial all the entities and relations following a uniform distribution. Each time we pick a triplet ($h, r, t$) from $\Delta$, an accompanied triplet ($h', r, t'$) is sampled at the same time by replacing the head or the tail with another entity from the entity set $E$, i.e.
$\Delta'_{(h, r, t)} = \{(h', r, t)|h'\in E\} \cup \{(h, r, t')|t'\in E\}$.
Then we choose one of the pair-wise SGD-based updating options depending on which camp ($\Delta^{+}_{(h, r, t)}$ or $\Delta^{-}_{(h,r,t)}$) that ($h', r, t'$) belongs to.

\begin{algorithm}
\caption{~~~~~The Learning Algorithm of {\bf LMNNE}}
\begin{algorithmic}[1]
\REQUIRE ~~\\
Training set $\Delta = \{(h, r, t)\}$, entity set $E$, relation set $R$;
dimension of embeddings $d$, margin $\gamma$, learning rate $\alpha$ and $\beta$ for $\mathcal{L}_{pull}$ and $\mathcal{L}_{push}$ respectively, convergence threshold $\epsilon$, maximum epoches $n$ and the trade-off $\mu$.\\

\FOR{${\bf r} \in R$}
\STATE ${\bf r} := \text {Uniform} (\frac{-6}{\sqrt{d}}, \frac{6}{\sqrt{d}})$
\STATE ${\bf r} :=  \frac{{\bf r}}{|{\bf r}|} $\
\ENDFOR

\STATE $i := 0$
\WHILE{$Rel. loss > \epsilon$ and $i < n$ }
\FOR{${\bf e} \in E$}
\STATE ${\bf e} := \text {Uniform}(\frac{-6}{\sqrt{d}}, \frac{6}{\sqrt{d}})$
\STATE ${\bf e} :=  \frac{{\bf e}}{|{\bf e}|} $\
\ENDFOR

\FOR{$(h, r, t) \in \Delta$}
\STATE $(h', r, t') := \text {Sampling} (\Delta'_{(h, r, t)})$
\IF {$(h', r, t') \in \Delta^{+}_{(h,r,t)}$}
\STATE $\text {Updating: } {\bf \bigtriangledown}_{(h, r, t, h', t')} \mathcal{L}_{pull} \text { with: } \alpha\mu$
\ENDIF
\IF {$(h', r, t') \in \Delta^{-}_{(h,r,t)}$}
\STATE $\text {Updating}  : {\bf \bigtriangledown}_{(h, r, t, h', t')} \mathcal{L}_{push} \text { with: } \beta(1-\mu)$
\ENDIF
\ENDFOR

\ENDWHILE
\\
\ENSURE ~~\\
All the embeddings of {\it e} and {\it r}, where ${\it e} \in E$ and ${\it r} \in R$.
\end{algorithmic}
\end{algorithm}

\section{Quantitative Experiments}
Embedding the knowledge into low-dimensional vector spaces makes it much easier for AI-related computing tasks, such as {\it Link prediction} (predicting $t$ given $h$ and $r$ or $h$ given $r$ and $t$) and {\it Triplet classification} (to discriminate whether a triplet $(h, r, t)$ is correct or wrong). Two latest studies \cite{Bordes2013a,Socher2013} used subsets of WordNet ({\bf WN}) and Freebase ({\bf FB}) data to evaluate their models and reported the performance on the two tasks respectively.

In order to conduct solid experiments, we compare our model ({\bf LMNNE}) with many related studies including state-of-the-art and baseline approaches involving in the two tasks, i.e. {\it Link prediction} and {\it Triplet classification}. All the datasets, the source codes and the learnt embeddings for entities and relations can be downloaded from \url{http://1drv.ms/1pUPwzP}.

\subsection{Link prediction}
One of the benefits on knowledge embedding is that we can apply simple vector calculations to many reasoning tasks. For example, {\it Link prediction} is a valuable task that contributes to completing the knowledge graph. Specifically, It aims at predicting the missing entity or the relation given the other two elements in a mutilated triplet.

With the help of knowledge embeddings, if we would like to tell whether the entity $h$ has the relation $r$ with the entity $t$, we just need to calculate the distance between ${\bf h + r}$ and ${\bf t}$. The closer they are, the more possibility the triplet ($h, r, t$) exists.
\subsubsection{Datasets}
Bordes et al. \cite{Bordes2013,Bordes2013a} released two benchmark datasets\footnote{The datasets can be downloaded from \url{https://www.hds.utc.fr/everest/doku.php?id=en:transe}} which were extracted from WordNet ({\bf WN18}) and Freebase ({\bf FB15K}). Table 1 shows the statistics of these two datasets. The scale of {\bf FB15K} dataset is larger than {\bf WN18} with much more relations but fewer entities.
\begin{table}[!htp]
\centering
\caption{Statistics of the datasets used for link prediction task.}
\begin{tabular}{|c|c|c|c|}
  \hline
  {\bf DATASET} & {\bf WN18} & {\bf FB15K}  \\
  \hline
  \hline
  \#(ENTITIES) & 40,943 & 14,951  \\
  \#(RELATIONS) & 18 & 1,345  \\
  \#(TRAINING EX.) & 141,442 & 483,142  \\
  \#(VALIDATING EX.) & 5,000 & 50,000  \\
  \#(TESTING EX.) & 5,000 & 59,071  \\
  \hline
\end{tabular}

\end{table}

\subsubsection{Evaluation Protocol}
For each testing triplet, all the other entities that appear in the training set take turns to replace the head entity. Then we get a bunch of candidate triplets associated with the testing triplet. The dissimilarity of each candidate triplet is firstly computed by the scoring functions, then sorted in ascending order, and finally the rank of the ground truth triplet is recorded. This whole procedure runs in the same way for replacing the tail entity, so that we can gain the mean results. We use two metrics, i.e. {\it Mean Rank} and {\it Mean Hit@10} (the proportion of ground truth triplets that rank in Top-10), to measure the performance. However, the results measured by those metrics are relatively inaccurate, as the procedure above tends to generate the false negative triplets. In other words, some of the candidate triplets rank rather higher than the ground truth triplet just because they also appear in the training set. We thus filter out those triplets to report more reasonable results.
\subsubsection{Experimental Results}
We compare our model {\bf LMNNE} with the state-of-the-art {\bf TransE} and other models mentioned in \cite{Bordes2013} and \cite{Bordes2014} evaluated on the {\bf WN18} and {\bf FB15K}. We tune the parameters of each previous model\footnote{All the codes for the related models can be downloaded from \url{https://github.com/glorotxa/SME}} based on the validation set, and select the combination of parameters which leads to the best performance. The results are almost the same as \cite{Bordes2013a}. For {\bf LMNNE}, we tried several combinations of parameters: $d = \{20, 50, 100\}$, $\gamma = \{0.1, 1.0, 2.0, 10.0\}$, $\alpha = \{0.01, 0.02, 0.05, 0.1, 0.5, 1.0\}$, $\beta = \{0.01, 0.02, 0.05, 0.1, 0.5, 1.0\}$ and $\mu = \{0.2, 0.4, 0.5, 0.6, 0.8\}$, and finally chose $d = 20$, $\gamma = 2.0$, $\alpha = \beta = 0.02$ and $\mu = 0.6$ for {\bf WN18} dataset; $d = 50$, $\gamma = 1.0$, $\alpha = \beta = 0.02$ and $\mu = 0.6$ for {\bf FB15K} dataset. Moreover, we adopted different distance metrics, such as $L_1$ {\it norm}, $L_2$ {\it norm} and {\it inner product}, for the scoring function. Experiments show that $L_2$ {\it norm} and $L_1$ {\it norm} are the best choices to measure the distances in $\mathcal{L}_{pull}$ and $\mathcal{L}_{push}$ for both of the two datasets, respectively. Table 2 demonstrates that {\bf LMNNE} outperforms all the prior arts, including the baseline model {\bf Unstructured} \cite{Bordes2014}, {\bf RESCAL} \cite{Nickel2011}, {\bf SE} \cite{Bordes2011}, {\bf SME (LINEAR)} \cite{Bordes2014}, {\bf SME (BILINEAR)} \cite{Bordes2014}, {\bf LFM} \cite{Jenatton2012} and the state-of-the-art {\bf TransE}\footnote{To conduct fair comparisons, we re-implemented {\bf TransE} based on the SGD algorithm (not mini-batch SGD) and fed the same random embeddings for initializing both {\bf LMNNE} and {\bf TransE}. That's the reason why our experimental results of {\bf TransE} are slightly different from the original paper \cite{Bordes2013,Bordes2013a}.} \cite{Bordes2013,Bordes2013a}, measured by {\it Mean Rank} and {\it Mean Hit@10}.

\begin{table*}
  \centering
    \caption{Link prediction results. We compared our proposed LMNNE with the state-of-the-art method (TransE) and other prior arts.}
  \begin{tabular}{*{9}{|c|}}
    \hline
    {\bf DATASET} & \multicolumn{4}{|c|}{\bf WN18} & \multicolumn{4}{|c|}{\bf FB15K}\\
    \hline
    \hline
    {\multirow{2}*{\bf METRIC}}
    & \multicolumn{2}{|c|}{\em MEAN RANK} & \multicolumn{2}{|c|}{\em MEAN HIT@10}& \multicolumn{2}{|c|}{\em MEAN RANK} & \multicolumn{2}{|c|}{\em MEAN HIT@10}\\
   & {\em Raw} & {\em Filter} & {\em Raw} & {\em Filter}& {\em Raw} & {\em Filter} & {\em Raw} & {\em Filter}\\
   \hline
   \hline
   {\bf Unstructured} & 315 & 304 & 35.3\% & 38.2\% & 1,074 & 979 & 4.5\% & 6.3\% \\
   {\bf RESCAL} & 1,180 & 1,163 & 37.2\% & 52.8\% & 828 & 683 & 28.4\% & 44.1\% \\
   {\bf SE} & 1,011 & 985 & 68.5\% & 80.5\% & 273 & 162 & 28.8\% & 39.8\% \\
   {\bf SME (LINEAR)} & 545 & 533 & 65.1\% & 74.1\% & 274 & 154 & 30.7\% & 40.8\% \\
   {\bf SME (BILINEAR)} & 526 & 509 & 54.7\% & 61.3\% & 284 & 158 & 31.3\% & 41.3\% \\
   {\bf LFM} & 469 & 456 & 71.4\% & 81.6\% & 283 & 164 & 26.0\% & 33.1\% \\
   {\bf TransE} & 294.4 & 283.2 & 70.4\% & 80.2\% &  243.3 & 139.9 & 36.7\% & 44.3\%  \\
   \hline
   {\bf LMNNE}  & {\bf 257.3} & {\bf 245.4} & {\bf 73.7\%} & {\bf 84.1\%} & {\bf 221.2} & {\bf 107.4} & {\bf 38.3\%} & {\bf 48.2\%} \\
   \hline
  \end{tabular}

\end{table*}
\begin{table*}
  \centering
    \caption{Results of {\it Filter Hit@10} (in \%) on FB15K categorized by different mapping properties of facts (M. stands for MANY).}
  \begin{tabular}{*{9}{|c|}}
      \hline
    {\bf TASK} & \multicolumn{4}{|c|}{\em Predicting head} & \multicolumn{4}{|c|}{\em Predicting tail}\\
    \hline
    {\bf REL. Mapping}  & 1-TO-1 & 1-TO-M. & M.-TO-1 & M.-TO-M. & 1-TO-1 & 1-TO-M. & M.-TO-1 & M.-TO-M.\\
    \hline
    \hline
    {\bf Unstructured}  & 34.5\% & 2.5\% & 6.1\% & 6.6\% & 34.3\% & 4.2\% & 1.9\% & 6.6\% \\
    {\bf SE}  & 35.6\% & 62.6\% & 17.2\% & 37.5\% & 34.9\% & 14.6\% & 68.3\% & 41.3\% \\
    {\bf SME (LINEAR)}  & 35.1\% & 53.7\% & 19.0\% & 40.3\% & 32.7\% & 14.9\% & 61.6\% & 43.3\% \\
    {\bf SME (BILINEAR)}  & 30.9\% & 69.6\% & {\bf 19.9\%} & 38.6\% & 28.2\% & 13.1\% & 76.0\% & 41.8\% \\
   {\bf TransE}  & 43.7\% & 65.7\% & 18.2\% & {\bf 47.2\%} & 43.7\% & 19.7\% & 66.7\% & 50.0\% \\
   \hline
   {\bf LMNNE}  & {\bf 59.2\%} & {\bf 77.8\%} & 17.5\% & 45.5\% & {\bf 58.6\%} & {\bf 20.0\%} & {\bf 80.9\%} & {\bf 51.2\%} \\
   \hline

  \end{tabular}

\end{table*}
Moreover, we divide {\bf FB15K} into different categories (i.e. ONE-TO-ONE, ONE-TO-MANY, MANY-TO-ONE and MANY-TO-MANY) based on the mapping properties of facts. According to {\bf TransE} \cite{Bordes2013a}, we set $1.5$ as the threshold to discriminate ONE and MANY. For example, given a pair ({\it h, r}), if the average number of tails appearing in the dataset is upon 1.5, we can categorize the triplets involving in this relation $r$ into the ONE-TO-MANY class. We evaluate the performance of {\bf Filter Hit@10} metric on each category. Table 3 shows that {\bf LMNNE} performs best on most categories. The result proves that the proposed approach can not only maintain the characteristic of modeling the ONE-TO-ONE triplets, but also better handle the facts with multi-mapping properties.

\subsection{Triplet classification}
Triplet classification is another task proposed by Socher et al. \cite{Socher2013} which focuses on searching a relation-specific distance threshold $\sigma_r$ to tell whether a triplet ($h, r, t$) is plausible.
\subsubsection{Datasets}
Similar to Bordes et al. \cite{Bordes2013,Bordes2013a}, Socher et al. \cite{Socher2013} also constructed two standard datasets\footnote{Those datasets can be download from the website \url{http://www.socher.org/index.php}}, i.e. {\bf WN11} and {\bf FB13}, sampled from WordNet and Freebase. However, both of the datasets contain much fewer relations. Therefore, we build another dataset following the principle proposed by Socher et al. \cite{Socher2013} based on {\bf FB15K} which owns much more relations. The head or the tail entity can be randomly replaced with another one to produce a negative triplet, but in order to build much tough validation and testing datasets, the principle emphasizes that the picked entity should once appear at the same position. For example, {\it (Pablo Picaso, nationality, American)} is a potential negative example rather than the obvious irrational {\it (Pablo Picaso, nationality, Van Gogh)}, given a positive triplet {\it (Pablo Picaso, nationality, Spainish)}, as {\it American} and {\it Spainish} are more common as the tails of {\it nationality}. Table 4 shows the statistics of the standard datasets that we used for evaluating models on the triplet classification task.

\begin{table}
\centering
\caption{Statistics of the datasets used for triplet classification task.}
\begin{tabular}{|c|c|c|c|}
  \hline
  {\bf DATASET} & {\bf WN11} & {\bf FB13} & {\bf FB15K} \\
  \hline
  \hline
  \#(ENTITIES) & 38,696 & 75,043 & 14,951 \\
  \#(RELATIONS) & 11 & 13 & 1,345  \\
  \#(TRAINING EX.) & 112,581 & 316,232 & 483,142 \\
  \#(VALIDATING EX.) & 5,218 & 11,816 & 100,000 \\
  \#(TESTING EX.) & 21,088 & 47,466 & 118,142 \\
  \hline
\end{tabular}

\end{table}

\subsubsection{Evaluation Protocol}
The decision strategy for binary classification is simple: If the dissimilarity of a testing triplet ($h, r, t$) computed by $f_r(h, t)$ is below the relation-specific threshold $\sigma_r$, it is predicted as positive, otherwise negative. The relation-specific threshold $\sigma_r$ can be searched via maximizing the classification accuracy on the validation triplets which belong to the relation $r$.
\subsubsection{Experimental Results}
We use the best combination of parameter settings in the {\it Link prediction} task ($d = 20$, $\gamma = 2.0$, $\alpha = \beta = 0.02$ and $\mu = 0.6$ for {\bf WN11} dataset; $d = 50$, $\gamma = 1.0$, $\alpha = \beta = 0.02$ and $\mu = 0.6$ for both {\bf FB13} and {\bf FB15K} datasets) to generate the entity and relation embeddings, and learn the best classification threshold $\sigma_r$ for each relation $r$. Compared with the state-of-the-art, i.e. {\bf TransE} \cite{Bordes2013a,Bordes2013} and other prior arts, such as {\bf Distance Model} \cite{Bordes2011}, {\bf Hadamard Model} \cite{Bordes2012}, {\bf Single Layer Model} \cite{Socher2013}, {\bf Bilinear Model} \cite{Sutskever2009,Jenatton2012} and {\bf Neural Tensor Network (NTN)}\footnote{Socher et al. reported higher classification accuracy in \cite{Socher2013} with word embeddings. In order to conduct a fair comparison, the accuracy of {\bf NTN} reported in Table 5 is same with the EV (entity vectors) results in Figure 4 of \cite{Socher2013}.} and \cite{Socher2013}, the proposed {\bf LMNNE} still achieves better performance as shown in Table 5.
\begin{table}
\centering
  \caption{The accuracy of triplet classification compared with the state-of-the-art method (TransE) and other prior arts.}
\begin{tabular}{|c|c|c|c|}
  \hline
  {\bf DATASET} & {\bf WN11} & {\bf FB13} & {\bf FB15K}\\
  \hline
  \hline
  {\bf Distance Model} & 53.0\% & 75.2\% & -  \\
  {\bf Hadamard Model}  & 70.0\% & 63.7\% & -  \\
  {\bf Single Layer Model}  & 69.9\% & 85.3\% & - \\
  {\bf Bilinear Model}  & 73.8\% &  84.3\% & - \\
  {\bf NTN} & 70.4\% &{\bf 87.1\%} & 66.7\% \\
  {\bf TransE} & 77.5\% &67.5\% & 85.8\% \\
  \hline
  {\bf LMNNE}& {\bf 78.6\%} & 74.8\% & {\bf 86.8\%}\\
  \hline
\end{tabular}

\end{table}

\begin{table*}
  \centering
  \caption{The scoring function and parameter complexity analysis for all the models mentioned in the experiments. For all the models, we assume that there are a total of $n_e$ entities, $n_r$ relations (In most cases, $n_e \gg n_r$.), and each entity is embedded into a $d$-dimensional vector space, i.e ${\bf h, t} \in \mathbb{R}^{d}$. We also suppose that there are $s$ slices in a tensor for the neural-network related models, i.e {\it Single Layer Model} and {\it Neural Tensor Network}.}
\begin{tabular}{|c|c|c|}
  \hline
  {\bf Model} & {\bf Scoring Function} & {\bf Parameter Complexity}\\
  \hline
  \hline
    {\bf Unstructured} & $||{\bf h}-{\bf t}||$ & $n_e d$\\
  \hline
    {\bf Distance Model (SE)} &  \tabincell{c}{
    $||W_{rh}{\bf h} - W_{rt}{\bf t}||$; \\ $(W_{rh}, W_{rt}) \in \mathbb{R}^{d \times d}$} & $n_e d + 2 n_r d^2$\\

  \hline
    {\bf Single Layer Model}  &  \tabincell{c}{
    ${\bf u}^T_r \tanh(W_{rh}{\bf h} + W_{rt}{\bf t} + {\bf b}_r)$; \\ $(W_{rh}, W_{rt}) \in \mathbb{R}^{s \times d}$, $({\bf u_r}, {\bf b}_r) \in \mathbb{R}^{s} $}& $n_e d + 2n_r(sd + s)$ \\

  \hline
    {\bf Bilinear Model} & \tabincell{c}{${\bf h}^TW_r{\bf t}$;\\$W_r \in \mathbb{R}^{d \times d}$} & $n_ed + n_rd^2$\\

  \hline
    {\bf Neural Tensor Network (NTN)} & \tabincell{c} {${\bf u}^T_r \tanh({\bf h}^TW_r{\bf t} + W_{rh}{\bf h} + W_{rt}{\bf t} + {\bf b}_r)$; \\ $W_r \in \mathbb{R}^{d \times d \times s}, (W_{rh}, W_{rt}) \in \mathbb{R}^{s \times d}$, $({\bf u_r}, {\bf b}_r) \in \mathbb{R}^{s}$}
    & $n_e d + n_r(sd^2 + 2sd + 2s)$\\
  \hline
    {\bf TransE} and {\bf LMNNE}& \tabincell{c} { $||{\bf h} + {\bf r} - {\bf t}||$;\\ ${\bf r} \in \mathbb{R}^d$}& $n_e d + n_r d$\\
        \hline

\end{tabular}

\end{table*}

\section{Qualitative Analysis}
In addition to the quantitative experiments of evaluating the performance on {\it Link prediction} and {\it Triplet classification} with several benchmark datasets, we analytically compare the parameter complexity among the approaches that we have mentioned as well. Table 6 lists the theoretical costs on representing triplets $(h, r, t)$ based on the scoring functions of nearly all the classical models. Except for {\bf Unstructured}, {\bf TransE} and {\bf LMNNE}, the other approaches regard the relation $r$ as a transportation matrix serving for the entities $h$ and $t$. These models need more resources on storing and computing embeddings. {\bf Unstructured} costs least, but does not contain any information on relations. {\bf LMNNE} and {\bf TransE} embed both entities and relations into the low-dimensional vector spaces from different aspects of observations: {\bf LMNNE} regards a relation as the inner connections of word embedding calculations, and {\bf TransE} considers it as a kind of translation from one entity to another. Despite the varies angles of modeling, both of them are relatively efficient models for knowledge representation.

\section{Conclusion and Future Work}
Knowledge embedding is an alternative way of representing knowledge besides displaying in triplets, i.e. ($h, r, t$). Its essence is to learn a distributed representation for each entity and relation, to make the knowledge computable, e.g. ${\bf h + r \approx t}$.

To achieve higher quality of embeddings, we propose {\bf LMNNE}, a both effective and efficient model on learning a low-dimensional vector representation for each entity and relation in Knowledge Bases.
Some canonical tasks, such as {\it Link prediction} and {\it Triplet classification}, which were ever based on hand-made logic rules, can be truly facilitated by means of the simple vector calculation.
The results of extensive experiments on several benchmark datasets and complexity analysis show that our model can achieve higher performance without sacrificing efficiency.

In the future, we look forward to paralleling the algorithm which can encode a whole KB with billion of facts, such as Freebase and YAGO. Another direction is that we can apply this new way of {\it Knowledge Representation} on reinforcing some other related studies, such as {\it Relation Extraction} from free texts and {\it Open Question Answering}.
\section{Acknowledgments}
This work is supported by National Program on Key Basic Research Project (973 Program) under Grant 2013CB329304, National Science Foundation of China (NSFC) under Grant No.61373075. The first author conducted this research while he was a joint-supervision Ph.D. student in New York University. This paper is dedicated to all the members of the Proteus Project\footnote{http://nlp.cs.nyu.edu/index.shtml}, and thanks so much to your help.
\bibliographystyle{IEEEtran}

\end{document}